# Emotion Recognition from Microblog Managing Emoticon with Text and Classifying using 1D CNN

[1]Md. Ahsan Habib, [1]M. A. H. Akhand and [2]Md. Abdus Samad Kamal

[1]*Department of Computer Science & Engineering, Khulna University of Engineering & Technology, Bangladesh*
[2]*Graduate School of Science and Technology, Gunma University, Japan*

Corresponding Author:
M. A. H. Akhand
Department of Computer Science & Engineering, Khulna University of Engineering & Technology, Bangladesh
Email: akhand@cse.kuet.ac.bd

**Abstract:** Microblog, an online-based broadcast medium, is a widely used forum for people to share their thoughts and opinions. Recently, Emotion Recognition (ER) from microblogs is an inspiring research topic in diverse areas. In the machine learning domain, automatic emotion recognition from microblogs is a challenging task, especially, for better outcomes considering diverse content. Emoticon becomes very common in the text of microblogs as it reinforces the meaning of content. This study proposes an emotion recognition scheme considering both the texts and emoticons from microblog data. Emoticons are considered unique expressions of the users' emotions and can be changed by the proper emotional words. The succession of emoticons appearing in the microblog data is preserved and a 1D Convolutional Neural Network (CNN) is employed for emotion classification. The experimental result shows that the proposed emotion recognition scheme outperforms the other existing methods while tested on Twitter data.

## Introduction

Emotion often refers to a complex state of feeling such as happiness, joy, anger, disgust, fear, love, and hatred that occurs in physical and psychological changes and has an impact on one's thinking and actions. Emotions can have a significant impact on people's lives. People's emotions can be expressed in a variety of ways, including speech, facial expressions, bodily gestures, verbal expressions using text, and so on (Castellano *et al*., 2008; Murugappan *et al*., 2021; Singh *et al*., 2020). Different social media platforms like Facebook, Twitter, Instagram, WhatsApp, Sina Weibo, etc. have grown in popularity in recent years where people express their emotions and thoughts (Che *et al*., 2021; Wang *et al*., 2016). Users share millions of tweets and posts every day. Therefore, social media contents are the most prospective sources for understanding emotional states and human thoughts.

Microblog, an online-based broadcast medium, is a widely used forum for people to share their thoughts and opinions. In our daily lives, microblogs make communication easier. Twitter (Java *et al*., 2007), LinkedIn, Facebook, Instagram, Snapchat, Tumblr, WhatsApp, etc. are the most popular microblogs. Globally, there are 3.5 billion users on social media, according to 2019 social media projections which equate to approximately 45% of the current population and this figure is only increasing (Mohsin, 2020). Any post can be hit immediately by a large number of people through microblogs. People express their thoughts by sharing posts that reflect one's sentiments.

Emotion Recognition (ER) from microblog data is the most promising and challenging research finding in the field of information and communication technology. Unimodal, bimodal, and multimodal are the three possible categories of ER. Unimodal emotion recognition uses only one type of information, such as facial expression, text, or speech, whereas bimodal emotion recognition uses both speech and facial expression. This study introduces the unimodal emotion recognition method to determine one's Emotion Category (EC). Machine Learning (ML) based approach and knowledge-based approach (Chaffar and Inkpen, 2011) are two major techniques for identifying or recognizing emotions. The knowledge-based technique also known as the lexicon-based technique uses a set of rules to detect emotion from given data (Nirenburg and Mahesh, 1997) whereas the ML-based technique uses a model for learning the patterns from features generated from microblog data (Kotsiantis *et al*., 2006).

Deep Learning (DL) based approaches for emotion recognition from microblog data have emerged remarkably and shown promising results nowadays. DL employs different architectures to learn the patterns in data. There are two major steps in DL-based emotion recognition from microblog data: (i) Processing data and (ii) classifying them using the proper DL model. Firstly, the collected data is processed, transformed, and represented in the appropriate format for the envisioned DL model. Secondly, a DL model is prepared or trained with the data to classify emotion. Along with different processing techniques, different DL models are investigated in the last several years for emotion recognition (Batbaatar *et al*., 2019; Guo *et al*., 2021a). These DL-based methods run on preprocessed data and do not include explicit functionality (Batbaatar *et al*., 2019; Yang *et al*., 2018). Some experiments consider emoticons along with the text as input data, while others only consider the text.

This research aims to develop an improved CNN based emotion recognition model from microblogs considering both text and emoticons. This study takes into account emoticons as they have a substantial significance to users' emotions and represented them by corresponding emotional words. The word and emoticon sequences that appeared in the microblog are preserved in this study. Considering all other pre-processing steps, CNN is applied for emotion recognition since it is pertinent for sequential data classification. Experiments using Twitter data on texts with emoticons and texts-only showed the effectiveness of the proposed CNN approach, with better classification accuracy considering both emoticons and texts compared to classification accuracy when only considering texts.

The remaining paper is structured as follows. At first, it reviews different existing works related to emotion recognition, and then the proposed CNN method is explained. After that, the experimental studies and results are demonstrated. Lastly, the paper concludes with a few remarks.

## Literature Review

Many DL-based approaches are examined in the last several years for emotion recognition from microblog data (Islam *et al*., 2020; Mehta *et al*., 2021). Attention model (Wei *et al*., 2019; Yuan and Zhang, 2021), BERT RCNN (Pan and Xu, 2021), CNN (Xu *et al*., 2020), GRU (Liu *et al*., 2021), LSTM (Arun *et al*., 2019; Batbaatar *et al*., 2019; Guo *et al*., 2021b; Islam *et al*., 2021), graph convolution network (Lai *et al*. 2020), etc., are most prominent techniques employed in this research domain. Most of the methods only consider textual data (Xu *et al*., 2020) and a few consider both text and emoticons (Islam *et al*., 2020).

Yang *et al*. (2018) developed an enhanced CNN method considering both emoticons and texts. They represented the emoticons and words as two separate vectors and projected into one emotional space. Then CNN is employed for emotion classification. The proposed model is applied to the Twitter dataset, NLPCC2013, and Weibo dataset.

The work proposed by Islam *et al*. (2020) also took both the emoticons and texts as input where it applied LSTM to classify emotions. The Twitter dataset was used to measure the efficacy of the model; however, the dataset was small enough. The work was then extended by Islam *et al*. (2021) and achieved remarkable accuracy with a relatively large dataset.

Batbaatar *et al*. (2019) introduced a Semantic Emotion Neural Network (SENN) model that includes both CNN and BiLSTM for ER. Here, the CNN model focused on the emotional connectivity between words after extracting emotional features, while the BiLSTM was employed to build the semantic relationship after collecting contextual information. The SENN used Twitter data and other social media data (only text) without specifying whether or not emoticons were used in the decision-making process.

Wei *et al*. (2019) developed an emotion recognition approach by incorporating both the dual attention mechanism and Bidirectional Long Short-term Memory (BiLSTM). In their work, they first used the BiLSTM model to semantically encode the microblog data and then introduced the sentiment word attention and self-attention into the BiLSTM model. Lastly, they used the Softmax classifier to classify the sentiment of microblogs. Chinese microblog Sina Weibo-based NLPCC2013 and NLPCC2014 datasets are used for the experimental purpose.

Another attention-based dual-channel microblog emotion recognition model is developed by Yuan and Zhang (2021). This study used RoBERTa-WWM and the multi-head attention model for their work. They constructed the emotional knowledge set of each sentence extending the emotional resource library and used the pre-training model RoBERTa-WWM for feature representation. After that, the Text CNN-BiGRU network and a Multi-Head Attention network took the sentence feature and the emotional knowledge as input to obtain deeper semantics features and attention features of emotional knowledge. And finally, the semantic feature and the emotional knowledge attention feature are combined to train the model. Chinese microblog NLPCC2014 dataset is used to show the efficacy of the model.

Pan and Xu (2021) developed a deep learning-based sentiment analysis model employing BERT RCNN for netizens during public health emergencies. They used BERT which uses static masking for fine-tuning the input data and trained them into vectors to represent it. Then it took the trained vectors as the input features of the upstream model and learned the microblog data features through RCNN network. In contrast, the work proposed by Yuan and Zhang (2021) used dynamic masking incorporated in RoBERTa making the model more robust.

Another model, called Semantic Emoticon Emotion Recognition (SEER) (Liu *et al*., 2021), used both the attention mechanism and Bi-GRU (bidirectional gated recurrent unit) network to classify emotion. Then they constructed an emoticon distribution model to obtain the emotion vectors.

Arun *et al*. (2019) developed EPUSAMCNN (Emotion-Prediction Using Semantic Analysis Multi-Dimensional Convolutional Neural Network) model incorporating both the LSTM (Long Short Term Memory Networks) and MCNN (Multi-Dimensional Convolutional Neural Network). They used MCNN to increase their proficiency in recognizing correct feelings for microblog data and BiLSTM for classifying them.

Another model named ERNIE-BiLSTM is developed for sentiment classification by Guo *et al*. (2021a). At first, they used ERNIE (Knowledge Enhanced Semantic Representation) pretrained model for word featuring. It considered both the enhancement of the semantic representation of words and preserves the contextual information along with the polysemy of words also. After training through ERNIE, they used BiLSTM for sentiment classification. They experimented with their developed model with the Chinese microblog Sina Weibo based NLPCC2014 dataset.

An emotion classification model is developed in 2020 by Lai *et al*. (2020). They used syntax based GCN (Graph Convolution Network) model focusing on the diverse grammatical structures. The accuracy of the model is enhanced by a percentile-based pooling technique proposed by them. They experimented with their developed model with the Chinese microblog dataset on their own.

## Emotion Recognition from Microblog Managing Emoticon with Text

Social media has been the most common medium of venting feelings in the era of globalization (Gräbner *et al*., 2012; Guo *et al*., 2021b; Wu *et al*., 2020; Xu *et al*., 2020). People share their thoughts by posting videos, texts, audio, photos, etc. to express emotions. Microblogs are the most popular among them. Millions of words, images, videos, audio, hashtags, and various signs and symbols with various meanings can be found on microblogs. One of the most widely used microblogging platforms is Twitter. Since emoticons reinforce the meaning of content, they should be given special consideration in emotion recognition along with texts.

The emoticon and its interactions with texts are given particular consideration in the proposed method. To identify the true emotion of people, both emoticons and text possess equal significance. Several pieces of research in the literature have described emoticons as noisy inputs that should be omitted during the pre-processing stage (Hogenboom *et al*., 2013) but this should not be. The proposed model did proper emotion analysis with the aid of emotional words and other texts in the microblog.

The working procedure of the proposed emotion recognition model is demonstrated in Fig. 1 for a sample post with an emoticon. The proposed CNN scheme contains four consecutive phases. A lookup table is used in Task 1 to translate emoticons into corresponding emotion words. In Task 2, Integer Encoding (IE), the Task of converting words into a series of integers, is done. Then in Task 3 padding is done to make an equal-length vector sequence of integers. Finally, CNN is applied to recognize specific emotions (Sad, Happy, Angry, or Love) in Task 4.

The whole methodology where four individual processes are shown is described in Algorithm 1. Data processing and CNN classification are two major tasks in this algorithm. The first three processing steps are under data processing. In the following subsections, the steps are described briefly.

**Algorithm 1:** Proposed ER scheme

```
Input: Microblog Data D of Word Size N
Output: Category of Emotion
// Task 1: Replacing emoticon(s) to corresponding meaning
For t = 1 to N do
    If (D[i] is emoticon(s)) then
        D[i] ← Emoticon. meaning (D[i])
    End If
End For
// Task 2: Integer Encoding (IE) using Tokenizer
For t = 1 to N do
    IE[i] = Tokenizer (D[i])
End For
// Task 3: Zero padding at first to make fixed L length
For t = 1 to L-N do
    P[i] ← 0 // Considering 0 for initial values
End For
For i = L-N+1 to N do
    P[i] ← IE[i] // Copy the rest values
End For
// Task 4: Emotion classification using CNN
// Embedding integer to 2D vector
For t = 1 to L do
    U [x, y] = Embedding (P[i])
End For
```

### *Data Processing*

One of the most critical phases in our developed scheme is to process microblog data. Among several social platforms' data, Twitter data is used including both emoticons and text. Certain pre-processing steps are required to eliminate unnecessary content and noisy input from the data. Case conversion, user name removal, hashtag, punctuation mark removal, and so on are all part of the cleaning process. Then the clean microblog data containing both texts and emoticons is processed into three processes. Task 1 (emoticon conversion step) searches microblog data for emoticon(s) and then uses the Emoticon.meaning() function emoticons are replaced with corresponding meaning. Individual emoticon word meanings are stored in a lookup table used by the function. Needless information (if any) is eliminated and IE is achieved in the Tokenization step (Task 2) using the function Text_to_sequence().

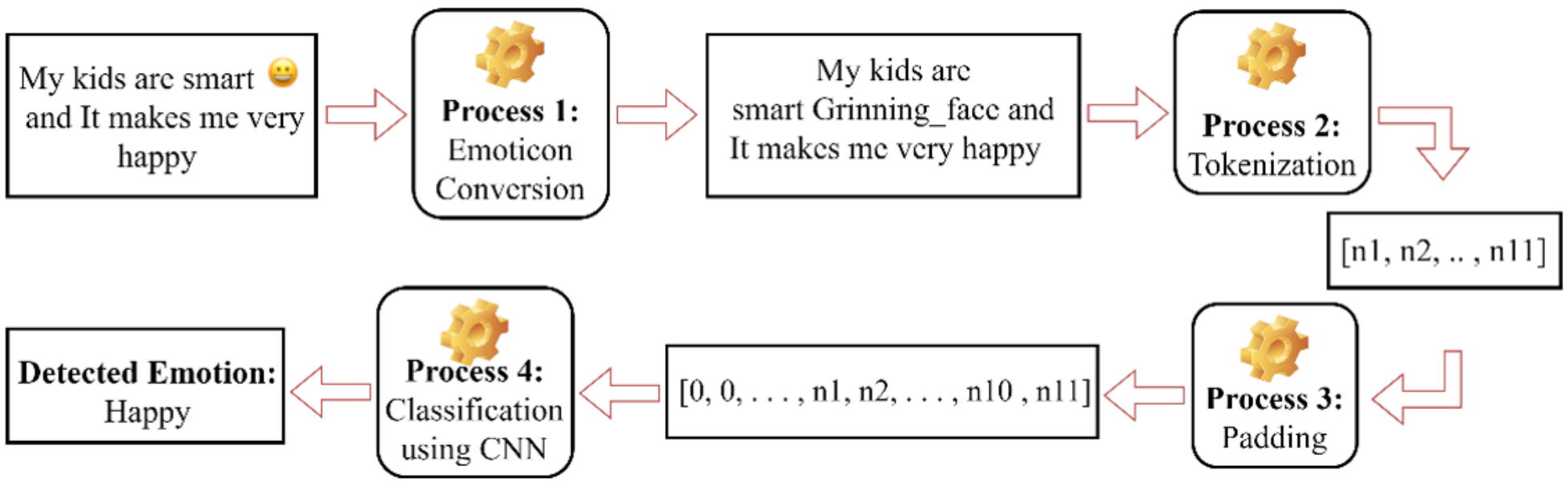

**Fig. 1:** The proposed ER architecture for a sample microblog data containing emoticons with texts

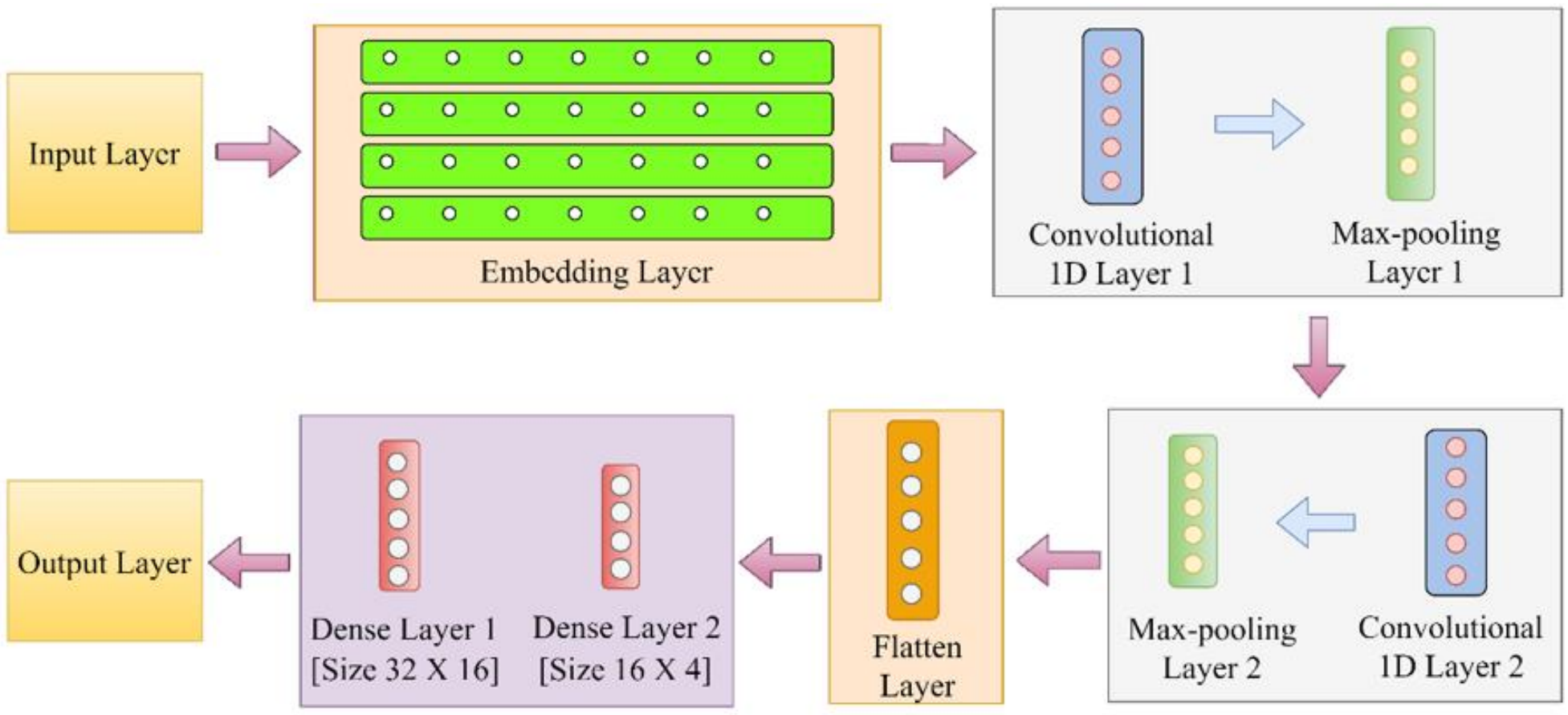

**Fig. 2:** 1D CNN architecture of the proposed ER

In Task 3, padding is done using the pad_sequence() function to form equal length word vector sequence. Zero padding at first is performed in this case. Lastly, in Task 4, the CNN model is applied to recognize particular emotions (Happy, Love, Sad, or Angry).

### *ER using 1D CNN*

CNN is a deep neural network very popular for analyzing visual 2D images. CNN has multiple layers having convolution and pooling operations. CNN's convolutional layers provide a summary of an image's features. By summarizing the presence of features of the feature map, pooling layers down samples feature maps. For images or image-like 2D inputs, the Conv2D layer architecture is primarily used in standard CNN. Finally, a fully connected layer is employed for classification purposes called Dense Layer.

This study considers CNN architecture with 1D convolutional operation on the blog text data mimicking the idea from CNN operation on time series data. For time-series data CNN uses Conv1D architecture (Amo-Boateng, 2020). As text data is considered time-series data CNN employs the Conv1D architecture for ER. The kernel in Conv1D slides in one dimensional way. Two Conv1D layers along with two max-pooling layers, a flattened layer, and two dense layers show promising outcomes in emotion recognition from text data. Generally, time-series data is used for forecasting or single output prediction but in our proposed method the output layer contains multiple nodes as emotion recognition is considered a multiclass problem.

Figure 2 illustrates the CNN architecture of the developed scheme for emotion recognition from microblog data. CNN is popular for analyzing texts and recognizing their features or patterns of them. The architecture of the proposed scheme consists of an input layer, an embedding layer, two convolutional layers, two maxpooling layers, a flatten layer, two dense layers, and lastly the output layer. The size of the words used in the proposed model is the input dimension of the embedding layer and the output dimension is 128. The first and second convolutional layers have 64 and 32 dimensionalities of output spaces, respectively. Both convolutional layers use kernel size 3 and the relu activation function. For both max-pooling layers, the max-pooling window size is set to 2 and the pooling window moves 2 strides for each pooling step. The following flattened layer flattens the input. The first dense layer contains 16 with a relu activation function. As there are four possible class labels, the proposed CNN architecture ends with a dense layer with four nodes. To get the probability for each class the softmax activation function is used.

## Results and Discussion

This segment demonstrates microblog data preparation, experimental setup, experimental results, and analysis of the proposed emotion recognition scheme.

### *Dataset Preparation*

This study uses the Twitter dataset, a pool of English tweets collected using the Twitter API. Firstly, tweeps, a python library for downloading tweets from Twitter, is used to collect tweets for this study purpose. The language filtering is enabled by the Twitter API, which permits the definition of the retrieved tweets' language. To extract English Tweets, the optional language parameter is set to 'en' in the Twitter Search URL. There are four emotion class labels in a total of 16011 tweets. The numeric notation of the class labels is as follows: Sad, happy, love, and angry are represented by 1, 2, 3, and 4 respectively. 75% (12008 tweets) of the collected data is used to train the proposed CNN architecture and the rest 25% (4003 tweets) is used as a test set in this study. Table 1 illustrates a snatch of the tweets with their corresponding EC. Table 2 displays the 16 emoticons that are used in the proposed scheme, along with their related word meaning.

### *Experimental Setup*

Keras (Powerful Open Source Python Library) text tokenization utility class is applied to convert the data words into numerical entities. Besides the 'Out of Vocabulary (OOV)' words can be handled here. The softmax and relu are considered activation functions for this emotion recognition multiclass classification problem. For the loss function and optimizer, the categorical-cross entropy and rmsprop are used, respectively. The proposed CNN model and data processing are implemented in the Python programming language. A web-based data-science environment such as "www.kaggle.com" is used to implement the experiment.

The proposed CNN approach is trained with batch sizes 32, 64, and 128 per batch. This experiment is run on a PC (Intel(R) Core (TM) i7-7700 CPU @ 3.60 GHz, RAM 16 GB, 64-bit OS) with Windows 10 environment OS.

### *Experimental Results and Performance Comparison*

The proposed CNN model's main benefit is that it considers emoticons in addition to emotion recognition from real-life Twitter data. Only text data is directed to the proposed CNN method without emoticons and the influence of emoticons in emotion recognition is observed. Figure 3 illustrates both training set and test set accuracies varying CNN training epochs up to 200 for several batch sizes. Compared to text-only accuracy, the proposed method achieves higher accuracy for both emoticon and text data as shown in the figure.

It is also worth mentioning that although the training set accuracy of text-only is consistent with the proposed CNN architecture, in terms of test accuracy, it obtains higher results than the text-only case. At batch size 128, the proposed CNN method achieves 39.9% test accuracy for the text-only data within 10 epochs. In contrast, the scheme achieves 88.0% test accuracy considering both emoticon and text at batch size 32 within 10 epochs. In any machine learning system, higher test set accuracy is desired because it is the indication of the system's ability for generalization. More accuracy in the test set specifies that emoticon use in addition to text boosted the capability of learning the emotion properly of the proposed CNN method.

**Table 1:** A Snatch of Tweets and Corresponding EC

| Microblog data (emoticons with texts) | Emotion category |
|---|---|
| Good morning 😄 | 2 |
| Today is not my day 😭 | 1 |
| 😭 I can't handle this | 1 |
| He looks ginger lol 😡 | 3 |
| 😡 I don't need the vaccine | 3 |
| Coming home to this period 😍 | 4 |

**Table 2:** Emoticons with corresponding word meanings

| Emoticon | Word meaning | Emoticon | Word meaning |
|---|---|---|---|
| 😀 | Grinning face | 😠 | Angry face |
| 😄 | Grinning face with smiling eyes | 😡 | Pouting face |
| 😁 | Beaming face with smiling eyes | 😤 | Face with steam from nose |
| ☺️ | Smiling face | 🤬 | Face with symbols on the mouth |
| 😭 | Loudly crying face | 😍 | Smiling face with heart-eyes |
| 😢 | Crying face | 🥰 | Smiling face with hearts |
| 🥺 | Pleading face | 😘 | Face blowing a kiss |
| ☹️ | Frowning face | 😚 | Kissing face with closed eyes |

Tables 3 and 4 demonstrate the confusion matrices of the developed emotion recognition model for emoticons with text and text-only cases, respectively. Figure 3(b) depicts the best test case accuracy for both cases. The differences between the predicted emotions and labeled emotions are shown in confusion matrices. These matrices depict four category-wise emotion recognition for emoticons with text and text-only cases. In terms of the test set, every category of emotion holds around 1000 tweets and the proposed CNN method accurately classifies 890 cases for the 'Angry' category which is the best performance using both text and emoticon data. The text-only case, on the other hand, performed best for the 'Sad' category, correctly classifying 429 out of 1000 cases. The other performance evaluation metric can be attained from the mentioned confusion matrices of the proposed CNN scheme for both cases.

**Table 3:** Confusion matrix of proposed CNN Scheme considering emoticon and text data

| | Predicted emotion category | | | | |
|---|---|---|---|---|---|
| Actual emotion category | Sad | Happy | Angry | Love | Total |
| Sad | 875 | 26 | 61 | 39 | 1001 |
| Happy | 31 | 882 | 45 | 43 | 1001 |
| Angry | 47 | 37 | 890 | 27 | 1001 |
| Love | 45 | 42 | 34 | 879 | 1000 |

**Table 4:** Confusion matrix for text data only

| | Predicted emotion category | | | | |
|---|---|---|---|---|---|
| Actual emotion category | Sad | Happy | Angry | Love | Total |
| Sad | 429 | 148 | 247 | 177 | 1001 |
| Happy | 187 | 412 | 208 | 194 | 1001 |
| Angry | 257 | 220 | 387 | 137 | 1001 |
| Love | 225 | 240 | 185 | 350 | 1000 |

**Table 5:** Comparison of the proposed CNN method with other existing methods on Twitter data

| Sl. | Work Ref, (Authors, Year) | Methodology | Sample size (Training + Test) | Test set accuracy |
|---|---|---|---|---|
| 1 | Wikarsa and Thahir (2015) | Naive Bayes | 268 (116+152) | 71.30% |
| 2 | Yang *et al*. (2018) | CNN | 7200 (4600+2600) | 72.60% |
| 3 | Batbaatar *et al*. (2019) | CNN + BiLSTM | 19,678 (Not Mentioned) | 61.30% |
| 4 | Islam *et al*. (2020) | LSTM | 3085 (2313+772) | 82.10% |
| 5 | Liu *et al*. (2021) | Bi-GRU | 10,000 (8000+2000) | 85.76% |
| | | | 15,000 (12,000+3000) | 86.35% |
| 6 | Proposed Method | CNN | 16,012 (12,009 + 4003) | 88.00% |

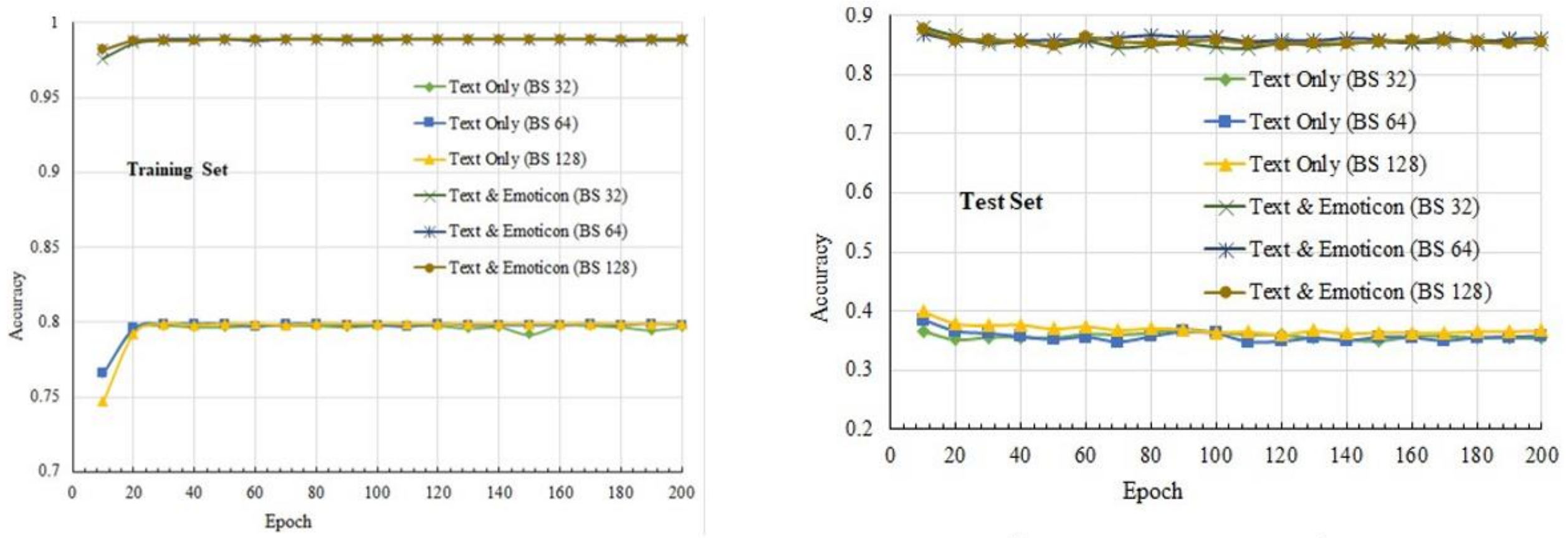

(a) Training Set Accuracy vs. Epochs
(b) Test Set Accuracy vs. Epochs

**Fig. 3:** Performance of ER for proposed CNN method (considering both emoticon and text) and for text-only for different Batch Sizes (BS)

Table 5 compares the proposed CNN method's emotion classification accuracy to other existing methods using Twitter data. The table also includes the methodology used in various studies and dataset sizes. The proposed method contains 16012 samples where 12009 tweets are for training and the rest 4003 tweets are for testing purposes. Wikarsa and Thahir (2015) used 268 tweets only 116 of them were used for training, which is not feasible. They applied Naive Bayes algorithm for classification and achieved an accuracy of 72.3%. Yang *et al*. (2018) used 4600 tweets for training purposes and applied CNN architecture to their model. Their model achieved 72.6% accuracy. Batbaatar *et al.* (2019) used 19,678 tweets for their study but didn't mention the train-test split ratio. They employed both CNN and BiLSTM and achieved an accuracy of 61.3%. For training purposes, Islam *et al*. (2020) used 2313 tweets which is not promising. They used LSTM based model and achieved 82.1% accuracy. On the other hand, Liu *et al.* (2021) employed two Chinese microblogs Sina Weibo based NLPCC2013 and NLPCC2014 datasets containing 10,000 and 15,000 sentences respectively. The model employed Bi-GRU architecture and achieved 85.76% accuracy for the NLPCC2013 dataset and 86.35% accuracy for the NLPCC2014 dataset. However, both datasets hold only 5400 sentences containing emoticons which might not produce a significant effect on classification. However, the proposed emotion recognition model with an 88.0% test set accuracy outperformed all other methods except (Islam *et al*., 2021). Showing competitive performance with (Islam *et al*., 2021), the proposed method has a significant contribution. For the emotion recognition model, CNN uses Conv1D architecture rather than the standard CNN Conv2D architecture that is most commonly used for imagery data. The proposed method successfully classifies emotion using CNN Conv1D architecture because text data is considered time series data. Finally, classification with CNN considering emoticons with texts has been revealed as a promising ER technique from microblog data.

## Conclusion

Nowadays, social media has become the most prevalent podium to express one's feelings & emotions and for a better kind, emoticons are commonly used with texts. In the ML domain, emotion recognition from microblog data has emerged as a challenging and promising research finding. For emotion recognition, most of the existing methods consider only text data for simplicity which is not sufficient. In this study, an emotion recognition model using CNN is developed considering emoticons in addition to text. As emoticons can have a significant role in the emotional behavior of human beings using microblog data, the proposed CNN technique outperforms other emotion recognition methods considering emoticons in addition to text using real-life Twitter data. In summary, this research developed an emotion classification technique and the effectiveness of emoticon consideration in emotion recognition from microblog data.

This study opens some future research scopes in this area. This study added a significant wing in the field of emotion recognition from English microblogs. A similar concept may be suitable for other language microblogs. Moreover, the recognition system brings more realistic if emotional states like surprise and disgust can be considered, which remained for further research. Moreover, with a large dataset, the proposed CNN approach could produce a more realistic result.

## Acknowledgment

The authors are thankful to Juyana Islam, Sr. Software Engineer at Samsung R&D Institute Bangladesh for sharing twitter dataset of their pilot study on emotion recognition (Islam *et al.* 2021).

## Funding Information

This research received no specific grant from any funding agency in the public, commercial, or not-for-profit sectors.

## Author’s Contributions

**Md. Ahsan Habib**: Participated in design, conducted experiments, performed result analysis and contributed to the writing of the manuscript.

**M. A. H. Akhand**: Designed the research plan and organized the study, analyzed and interpreted results and prepared the manuscript.

**M. A. S. Kamal**: Participated in design, contributed to model illustration and reviewed the manuscript.

## Ethics

It has been testified by the authors that this article has not been submitted to be published in any other journal and contains no ethical issues.

## References

Arun, V., Vineeth, R., & Prudhvi, C. (2019). Emotion prediction using semantic analysis neural network. *Journal of Advanced Research in Dynamical and Control Systems*, 11(4), 1184-1191.

Amo-Boateng, M. (2020). Tracking and Classifying Global COVID-19 Cases by using 1D Deep Convolution Neural Networks. *medRxiv*. https://doi.org/10.1101/2020.06.09.20126565

Batbaatar, E., Li, M., & Ryu, K. H. (2019). Semantic-emotion neural network for emotion recognition from text. *IEEE Access*, *7*, 111866-111878. https://ieeexplore.ieee.org/abstract/document/8794541/

Castellano, G., Kessous, L., & Caridakis, G. (2008). Emotion recognition through multiple modalities: Face, body gesture, speech. In *Affect and emotion in human-computer interaction* (pp. 92-103). Springer, Berlin, Heidelberg. https://doi.org/10.1007/978-3-540-85099-1_8

Chaffar, S., & Inkpen, D. (2011, May). Using a heterogeneous dataset for emotion analysis in the text. In *Canadian conference on artificial intelligence* (pp. 62-67). Springer, Berlin, Heidelberg. https://doi.org/10.1007/978-3-642-21043-3_8

Che, Z., Chang, J., Zhang, H., & Du, F. (2021, September). A Microblog Popularity Prediction Model Based on Temporal Sequence Features and Text Features. In *2021 IEEE International Conference on Computer Science, Electronic Information Engineering and Intelligent Control Technology (CEI)* (pp. 795-800). IEEE. https://ieeexplore.ieee.org/abstract/document/9574577

Gräbner, D., Zanker, M., Fliedl, G., & Fuchs, M. (2012, January). Classification of customer reviews based on sentiment analysis. In *ENTER* (pp. 460-470).

Guo, H., Chi, C., & Zhan, X. (2021a, June). ERNIE-BiLSTM Based Chinese Text Sentiment Classification Method. In *2021 International Conference on Computer Engineering and Application (ICCEA)* (pp. 84-88). IEEE. https://ieeexplore.ieee.org/abstract/document/9581123

Guo, X., Lai, H. *et al*. (2021b). "Emotion Classification of Case-Related Microblog Comments Integrating Emotional Knowledge." *Jisuanji Xuebao/Chinese Journal of Computers, 44(3)*, 564-578.

Hogenboom, A., Bal, D., Frasincar, F., Bal, M., de Jong, F., & Kaymak, U. (2013, March). Exploiting emoticons in sentiment analysis. In *Proceedings of the 28th annual ACM symposium on applied computing* (pp. 703-710). https://doi.org/10.1145/2480362.2480498

Islam, J., Ahmed, S., Akhand, M. A. H., & Siddique, N. (2020, June). Improved Emotion Recognition from Microblog Focusing on Both Emoticon and Text. In *2020 IEEE Region 10 Symposium (TENSYMP)* (pp. 778-782). IEEE. https://ieeexplore.ieee.org/abstract/document/9230725

Islam, J., Akhand, M. A. H., Habib, M. A., Kamal, M. A. S., & Siddique, N. (2021). Recognition of Emotion from Emoticon with Text in Microblog Using LSTM. *Advances in Science, Technology and Engineering Systems Journal*, *6*(3), 347-354. https://pdfs.semanticscholar.org/0733/87af99f576142723b3340203d80d86532f3a.pdf

Java, A., Song, X., Finin, T., & Tseng, B. (2007, August). Why we twitter: Understanding microblogging usage and communities. In *Proceedings of the 9th WebKDD and 1st SNA-KDD 2007 workshop on Web mining and social network analysis* (pp. 56-65). https://doi.org/10.1145/1348549.1348556

Kotsiantis, S. B., Zaharakis, I. D., & Pintelas, P. E. (2006). Machine learning: A review of classification and combining techniques. *Artificial Intelligence Review*, *26*(3), 159-190. https://doi.org/10.1007/s10462-007-9052-3

Lai, Y., Zhang, L., Han, D., Zhou, R., & Wang, G. (2020). Fine-grained emotion classification of Chinese microblogs based on graph convolution networks. *World Wide Web*, *23*(5), 2771-2787. https://doi.org/10.1007/s11280-020-00803-0

Liu, C., Liu, T., Yang, S., & Du, Y. (2021). Individual Emotion Recognition Approach Combined Gated Recurrent Unit with Emoticon Distribution Model. *IEEE Access*, *9*, 163542-163553. https://ieeexplore.ieee.org/abstract/document/9597507

Mehta, A., Patil, S., Dave S., & Tawde P. (2021). "Emotion Detection Using Social Media Data" *International Journal for Research in Applied Science and Engineering Technology* 9(11), 1456-1459. https://doi.org/10.22214/ijraset.2021.39027

Mohsin, M. (2020). "10 Social media statistics you need to know in 2021 [Infographic]." Oberlo. https://www.oberlo.com/blog/social-media-marketing-statistics

Murugappan, M., Zheng, B. S., & Khairunizam, W. (2021). Recurrent quantification analysis-based emotion classification in stroke using electroencephalogram signals. *Arabian Journal for Science and Engineering*, *46*(10), 9573-9588. https://doi.org/10.1007/s13369-021-05369-1

Nirenburg, S., & Mahesh, K. (1997). Knowledge-Based Systems for Natural Language Processing. *The Computer Science and Engineering Handbook*, *1997*, 637-653.

Pan, Z., & Xu, W. (2021, August). Deep learning based sentiment analysis during public health emergency. In *2021 13th International Conference on Intelligent Human-Machine Systems and Cybernetics (IHMSC)* (pp. 137-140). IEEE. https://ieeexplore.ieee.org/abstract/document/9555964

Singh, R., Puri, H., Aggarwal, N., & Gupta, V. (2020). An efficient language-independent acoustic emotion classification system. *Arabian Journal for Science and Engineering*, *45*(4), 3111-3121. https://link.springer.com/article/10.1007/s13369-019-04293-9

Wang, L., Wang, M., Guo, X., & Qin, X. (2016). Microblog sentiment orientation detection using user interactive relationship. *Journal of Electrical and Computer Engineering*, *2016*. https://doi.org/10.1155/2016/7282913

Wei, W., Zhang, Y., Duan, R., & Zhang, W. (2019, June). Microblog sentiment classification method based on dual attention mechanism and bidirectional LSTM. In *Workshop on Chinese Lexical Semantics* (pp. 309-320). Springer, Cham. https://doi.org/10.1007/978-3-030-38189-9_33

Wikarsa, L., & Thahir, S. N. (2015, November). A text mining application of emotion classifications of Twitter's users using Naive Bayes method. In *2015 1st International Conference on Wireless and Telematics (ICWT)* (pp. 1-6). IEEE. https://ieeexplore.ieee.org/abstract/document/7449218

Wu, P., Li, X., Shen, S., & He, D. (2020). Social media opinion summarization using emotion cognition and convolutional neural networks. *International Journal of Information Management*, *51*, 101978. https://doi.org/10.1016/j.ijinfomgt.2019.07.004

Xu, D., Tian, Z., Lai, R., Kong, X., Tan, Z., & Shi, W. (2020). Deep learning based emotion analysis of microblog texts. *Information Fusion*, *64*, 1-11. https://doi.org/10.1016/j.inffus.2020.06.002

Yang, G., He, H., & Chen, Q. (2018). Emotion-semantic-enhanced neural network. *IEEE/ACM Transactions on Audio, Speech and Language Processing*, *27*(3), 531-543. https://ieeexplore.ieee.org/abstract/document/8573804

Yuan, K., & Zhang, M. (2021, July). Dual-Channel Microblog Emotion Analysis Based on RoBERTa-WWM and Multi-Head Attention. In *The International Conference on Natural Computation, Fuzzy Systems and Knowledge Discovery* (pp. 486-495). Springer, Cham. https://doi.org/10.1007/978-3-030-89698-0_50